\documentclass[10pt,twocolumn,letterpaper]{article}

\usepackage{cvpr}
\usepackage{times}
\usepackage{epsfig}
\usepackage{graphicx}
\usepackage{amsmath}
\usepackage{amssymb}
\usepackage{subcaption}

\cvprfinalcopy 

\def\cvprPaperID{****} 

\begin{document}

\title{Object detection at 200 Frames Per Second }

\author{Rakesh Mehta\\
United Technologies Research Center-Ireland\\
{\tt\small mehtar1@utrc.utc.com}
\and
Cemalettin Ozturk \\
United Technologies Research Center-Ireland\\
{\tt\small OzturkC@utrc.utc.com}
}

\maketitle

\begin{abstract}
In this paper, we propose an efficient and fast object detector which can process hundreds of frames per second. To achieve this goal we investigate three main aspects of the object detection framework: network architecture,  loss function and training data (labeled and unlabeled). In order to obtain compact network architecture, we introduce various improvements, based on recent work, to develop an architecture which is computationally light-weight and achieves a reasonable performance. To further improve the performance, while keeping the complexity same, we utilize distillation loss function. Using distillation loss we transfer the knowledge of a more accurate teacher network to proposed light-weight student network.  We propose various innovations to make distillation efficient for the proposed one stage detector pipeline: objectness scaled distillation loss, feature map non-maximal suppression and a single unified distillation loss function for detection. Finally, building upon the distillation loss, we explore how much can we push the performance by utilizing the unlabeled data. We train our model with unlabeled data using the soft labels of the teacher network. Our final network consists of 10x fewer parameters than the VGG based object detection network and it achieves a speed of more than 200 FPS and proposed changes improve the detection accuracy by 14 mAP over the baseline on Pascal dataset.
\end{abstract}

\section{Introduction}
Object detection is a fundamental problem in computer vision. In recent years we have witnessed a significant improvement in the accuracy of object detectors \cite{redmon2016you,ren2015faster,liu2016ssd,redmon2016yolo9000, girshick2014rich,he2017mask,lin2017feature} owing to the success of deep convolutional networks \cite{krizhevsky2012imagenet}. It has been shown that modern deep learning based object detectors can detect a number of a generic object with considerably high accuracy and at a reasonable speed \cite{liu2016ssd,redmon2016yolo9000}. These developments have triggered the use of object detection in various industrial applications such as surveillance, autonomous driving, and robotics. Majority of the research in this domain has been focused towards achieving the state-of-the-art performance on the public benchmarks \cite{lin2017feature,he2017mask}. For these advancements the research has relied mostly on deeper architectures (Inception \cite{szegedy2014scalable},  VGG \cite{simonyan2014very}, Resnet \cite{he2016deep}) which come at the expense of a higher computational complexity and additional memory requirements. Although these results have demonstrated the applicability of object detection for a number of problems, the scalability still remains an open issue for full-scale industrial deployment. For instance, a security system with fifty cameras and a 30 frame/sec rate, would require a dedicated server with 60 GPUs even if we use one of the fastest detector SSD (22 FPS at 512 resolution) \cite{liu2016ssd}. These number can quickly grow for a number of industrial problems for example security application in a big building. In these scenarios, the speed and memory requirement becomes crucial as it enables processing of multiple streams on a single GPU. Surprisingly, the researchers have given little importance to the design of fast and efficient object detectors which have low memory requirements \cite{kim2016pvanet}. In this work we try to bridge this gap, we focus on the development of an efficient object detector with low memory requirements and a high speed to process multi-streams on a single GPU. 


  

\begin{figure}
\centering
   \includegraphics[trim = {4cm 11cm 4cm 10cm},clip, width=0.9\linewidth]{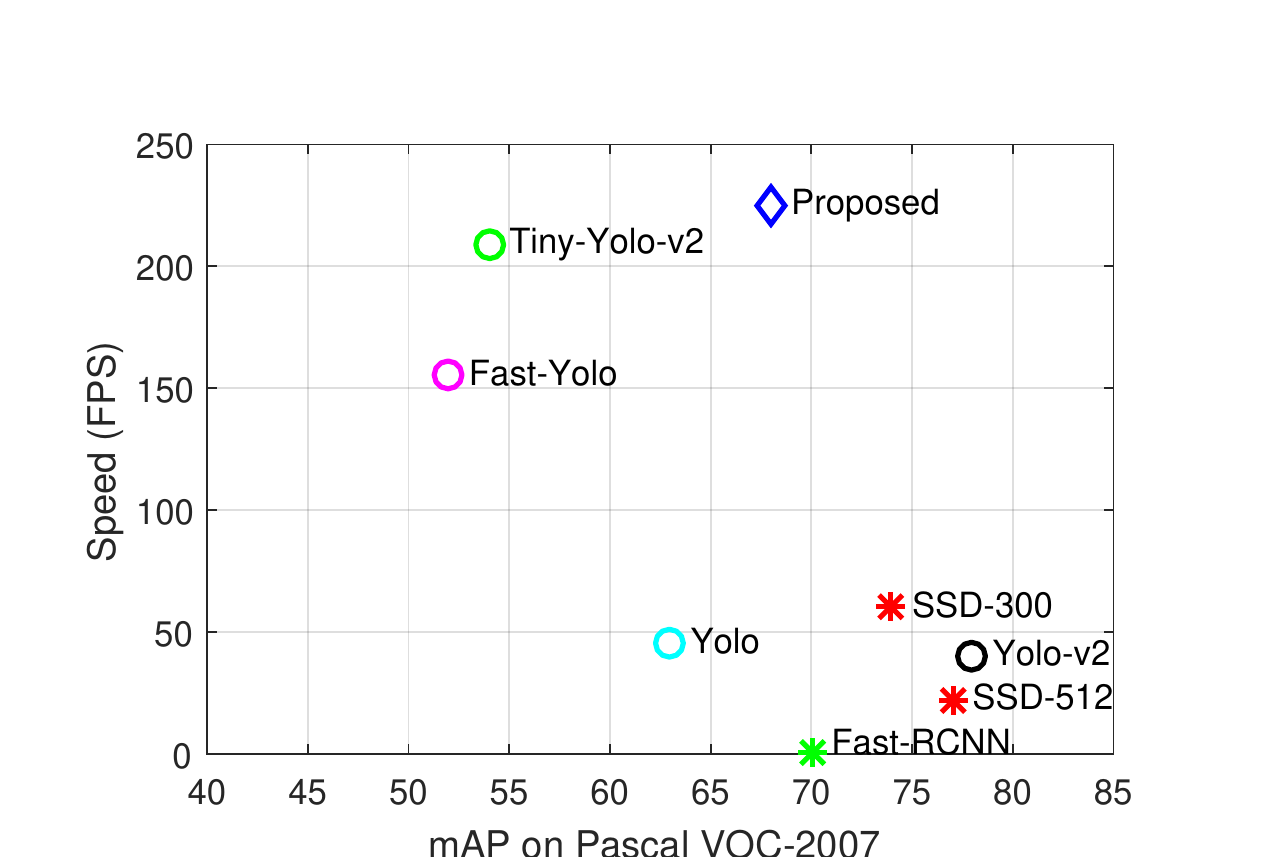}
  \caption{ Speed and performance comparison of the proposed detector with other competing approaches. For SSD and Yolo-v2 we show results for more accurate models. }
     \label{fig:map-speed-comparison}
\end{figure}

 With the aim to design a fast and efficient object detector we start by asking ourselves a fundamental question: what are the essential elements of a deep learning object detector and how can we tailor them to develop the envisaged detector? Based on the related work \cite{redmon2016you,ren2015faster,liu2016ssd,redmon2016yolo9000, girshick2014rich} we broadly identify the key components of the deep learning based object detection framework as (1) Network architecture, (2) Loss function and (3) Training data. We study each of these components separately and introduce a broad set of customizations, both novel and from related work, and investigate which of these play the most crucial role in achieving a speed-accuracy trade-off. 
 
 The network architecture is a key factor which determines the speed and the accuracy of the detector. Recent detectors \cite{huang2017speed} tends be based on deeper architecture (VGG \cite{simonyan2014very} , Resnet \cite{he2016deep}), which makes them accurate but increases the computationally complexity. Network compression and pruning \cite{han2015deep, denil2013predicting, howard2017mobilenets} approaches have been utilized with some success to make detection architectures more compact \cite{zhang2017shufflenet,shen2017dsod,zoph2017learning}. However, these approaches still struggle to improve the speed, for instance, the compact architecture of \cite{shen2017dsod} with 17M parameters achieves a speed of 17 FPS. In this work, we develop certain design principles not only for compact architecture but also for a high speed detector, as speed is of prime importance for us. We draw inspiration from the recent work of Densenet\cite{huang2017densely}, Yolo-v2 \cite{redmon2016yolo9000} and Single Shot Detector (SSD) \cite{liu2016ssd}, design an architecture which is deep but narrow. The deeper architecture allows us to achieve higher accuracy while the narrow layers enable us to control the complexity of the network. It is observed that the architectural changes itself can result in 5 mAP increase over the selected baseline. Building on these works, our main contribution in architecture design is a development of a simple yet efficient network architecture which can process more than 200 FPS making it the fastest deep learning based object detector.  Furthermore, our model contains only 15M parameters compared to 138M in VGG-16 model \cite{chen2017learning} thus resulting in one of the most compact networks.  The speed of the proposed detector in comparison to state-of-the-art detector is shown in Fig. \ref{fig:map-speed-comparison}.
 
  Given the restriction of simple and fast architecture, we investigate efficient training approaches to improve the performance. Starting with a reasonably accurate lightweight detector we leverage deeper networks with better performance to further improve the training strategy. For this purpose, we consider network distillation \cite{hinton2015distilling,buciluǎ2006model,ba2014deep}, where the knowledge of a larger network is utilized to efficiently learn the representation for the smaller network. Although the idea was applied to object detection recently \cite{chen2017learning, li2017mimicking}, our work has key contributions in the way we apply distillation. (1) We are the first one to apply distillation on single pass detector (Yolo), which makes this work different from the prior work which applies it to the region proposal network. (2) The key to our approach is based on the observation that object detection involves non-maximal suppression (NMS) step which is outside the end-to-end learning. Prior to NMS step, the last layer of the detection network consists of dense activations in the region of detection, if directly transferred to the student network it leads to over-fitting and deteriorates the performance. Therefore, in order to apply distillation for detection, we propose Feature Map-NMS (FM-NMS) which suppresses the activation corresponding to overlapping detections.  (3) We formulate the problem as an objectness scaled distillation loss by emphasizing the detections which have higher values of objectness in the teacher detection. Our results demonstrate the distillation is an efficient approach to improving the performance while keeping the complexity low.
  
Finally, we investigate ``the effectiveness of data'' \cite{halevy2009unreasonable} in the context of object detection. Annotated data is limited but with the availability of highly accurate object detectors and an unlimited amount of unlabeled data, we explore how much we can push the performance of the proposed light-weight detector. Our idea follows the line of semi-supervised learning \cite{rosenberg2005semi,yarowsky1995unsupervised, chen2013neil} which has not been thoroughly explored in deep learning object detector. Closely related to our approach is the recent work of Radosavovic et. al. \cite{radosavovic2017data} where annotations are generated using an ensemble of detectors. Our idea differs from their's in two main ways: (1) We transfer the soft labels from the convolutional feature maps of teacher network, which has shown to be more efficient in network distillation \cite{romero2014fitnets}. (2) Our loss formulation, through objectness scaling and distillation weight, allows us to control the weight given to the teacher label. This formulation provides the flexibility to give high importance to ground-truth detections and relatively less to inaccurate teacher prediction.  Furthermore, our training loss formulation seamlessly integrates the detection loss with the distillation loss, which enables the network to learn from the mixture of labeled and unlabeled data. To the best of our knowledge, this is the first work which trains the deep learning object detector by jointly using the labeled and unlabeled data.






\begin{figure}
\centering
   \includegraphics[trim = {1cm 12cm 16cm 2cm}, clip, width=0.95\linewidth]{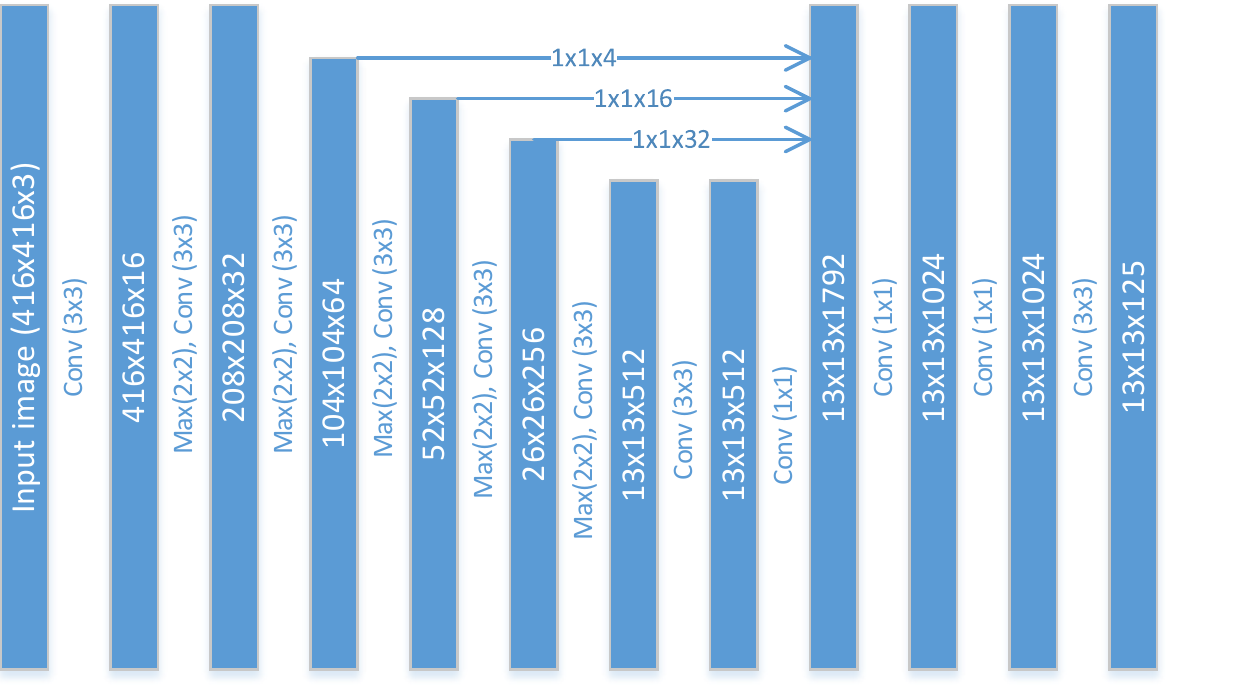}
  \caption{ Base architecture of our detector. To keep architecture simple we limit the depth of the network, keep number of feature maps low and use a small filter kernel ($3 \times 3$ or $1 \times 1$). }
     \label{fig:architecture}
\end{figure}

\section{Architecture customizations}
Most of the recent successful object detectors are dependent on the depth of the underlying architecture to achieve good performance. They achieve good performance, however, the speed is restricted to 20-60 FPS even for the fastest detectors\cite{liu2016ssd,redmon2016yolo9000} . In order to develop a much faster detector, we select a moderately accurate but a really high-speed detector, Tiny-Yolo \cite{redmon2016yolo9000}, as our baseline. It is a simplified version of Yolo-v2 with fewer convolutional layers, however with the same loss function and optimization strategies such as batch normalization \cite{ioffe2015batch}, dimension cluster anchor box, etc. Building upon this we introduce a number of architectural customizations to make it more accurate and even faster.  

\textbf{Dense feature map with stacking}
Taking inspiration from recent works \cite{huang2017densely,liu2016ssd} we observe that merging the feature maps from the previous layers improves the performance. We merge the feature maps from a number previous layer in the last major convolutional layer. The dimensions of the earlier layers are different from the more advanced one. Prior work \cite{liu2016ssd} utilizes max pooling to resize the feature maps for concatenation. However, we observe that the max pooling results in a loss of information, therefore, we use feature stacking where the larger feature maps are resized such that their activations are distributed along different feature maps \cite{redmon2016yolo9000}. 

Furthermore, we make extensive use of the bottleneck layers while merging the features. The idea behind the use of bottleneck layer \cite{huang2017densely, szegedy2016rethinking} is to compress the information in fewer layers. The 1x1 convolution layers in the bottleneck provide the flexibility to express the compression ratio while also adding depth at the same time. It is observed that merging the feature maps of advanced layers provide more improvement, therefore, we use a higher compression ratio for the initial layers and lower one for the more advanced layers.

\textbf{Deep but narrow}
The baseline Tiny-Yolo architecture consists of a large number (1024) of feature channels in their last few convolutional layers. With the feature map concatenation from the prior layers, we found the that these large number of convolutional features maps are not necessary. Therefore, the number of filters are reduced in our design, which help in improving the speed. 

Compared to other state-of-the-art detectors, our architecture lacks depth. Increasing depth by adding more convolutional layers results in computational overhead, in order to limit the complexity we use 1x1 convolutional layers. After the last major convolutional layer, we add a number of 1x1 convolutional layers which add depth to the network without increasing computational complexity.  

\textbf{Overall architecture}
Building on these simple concepts we develop our light-weight architecture for the object detection. These modification results in an improvement of 5 mAP over the baseline Tiny-Yolo architecture, furthermore, our architecture is 20\% faster than Tiny-Yolo because we use fewer convolutional filters in our design. The baseline Tiny-Yolo achieves a 54.2 mAP on Pascal dataset, while the proposed architecture obtains 59.4 mAP. The overall architecture is shown in Fig. \ref{fig:architecture}. We address this network as F-Yolo.

\begin{figure}
\centering
   \includegraphics[clip, width=\linewidth]{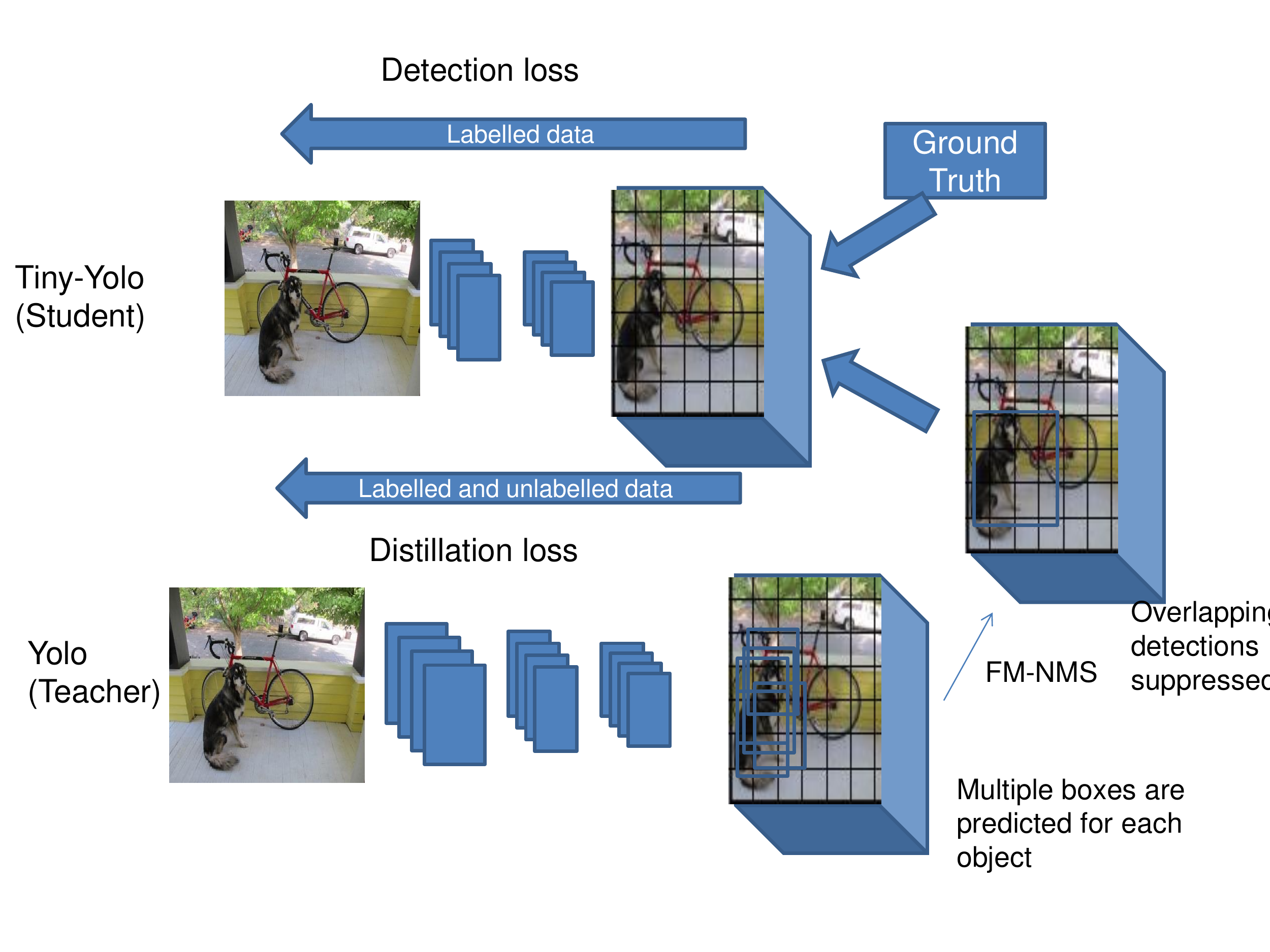}
  \caption{ Overall architecture for the distillation approach. Distillation loss is used for both labelled and unlabelled data. FM-NMS is applied on the last layer feature maps of teacher network to supress the overlapping candidates. }
     \label{fig:distillnms}
\end{figure}

\section{Distillation loss for Training}
Since we restrict ourselves to a simple architecture to achieve high speed, we explore network distillation \cite{romero2014fitnets,hinton2015distilling} approach to improve the performance as it does not affect the computational complexity of the detector. The idea of the network distillation is to train a light-weight (student) network using knowledge of a large accurate network (teacher). The knowledge is transferred in form of soft labels of the teacher network. 

 Before describing our distillation approach, we provide a brief overview of the Yolo loss function and the last convolutional layer of the network. Yolo is based on one stage architecture, therefore, unlike RCNN family of detectors, both the bounding box coordinates and the classification probabilities are predicted simultaneously as the output of the last layer. Each cell location in the last layer feature map predicts $N$ bounding boxes, where $N$ is the number of anchor boxes. Therefore,  the number of feature maps in the last layers are set to $N \times (K + 5)$, where K is the number of classes for prediction of class probabilities and 5 corresponds to bounding box coordinates and objectness values (4+1). Thus, for each anchor box and in each cell, network learns to predict: class probabilities, objectness values and bounding box coordinates. The overall objective can be decomposed into three parts: regression loss, objectness loss and classification loss. We denote the multi-part objective function as:
\begin{equation}
L_{Yolo} = f_{obj} (o_i^{gt}, \hat{o_i}) + f_{cl}(p_i^{gt}, \hat{p_i}) + f_{bb} (b_i^{gt}, \hat{b_i})
\end{equation}
where $\hat{o_i}, \hat{p_i}, \hat{b_i}$ are the objectness, class probability and bounding box coordinates of the student network and $o_i^{gt}, p_i^{gt}, b_i^{gt}$ are the values derived from the ground truth. The objectness is defined as IOU between prediction and ground truth, class probabilities are the conditional probability of a class given there is an object, the box coordinates are predicted relative to the image size and loss functions are simple $L_1$ or $L_2$ functions see \cite{redmon2016you,redmon2016yolo9000} for details. 

 To apply distillation we can simply take the output of the last layer of the teacher network and replace it with the ground truth $o_i^{gt}, p_i^{gt}, b_i^{gt}$. The loss would propagate the activations of the teacher network to student network. However, the dense sampling of the single stage detector introduces certain problems which makes the straightforward application of distillation ineffective. We discuss these problems below and provide simple solutions for applying distillation in single stage detector.


\subsection{Objectness scaled Distillation}
Current distillation approaches for detectors (applied for RCNN family) \cite{li2017mimicking,chen2017learning} use the output of the last convolutional layer of the teacher network for transferring the knowledge to the student network. Following similar approach in Yolo, we encounter a problem because it is single stage detector and predictions are made for a dense set of candidates. Yolo teacher predicts bounding boxes in the background regions of an image. During inference, these background boxes are ignored by considering the objectness value of the candidates. However, standard distillation approach transfers these background detections to the student. It impacts the bounding box training $f_{bb}()$, as the student network learns the erroneous bounding box in the background region predicted by the teacher network. The RCNN based object detectors circumvent this problem with the use of region proposal network which predicts relatively fewer region proposals. In order to avoid ``learning'' the teacher predictions for background region, we formulate the distillation loss as objectness scaled function. The idea is to learn the bounding box coordinates and class probabilities only when objectness value of the teacher prediction is high.  The objectness part of the function does not require objectness scaling because the objectness values for the noisy candidates are low, thereby the objectness part is given as:
\begin{equation}
f_{obj}^{Comb} (o_i^{gt},\hat{o_i}, o_i^{T}) = \underbrace{ f_{obj} (o_i^{gt}, \hat{o_i})}_\text{Detection loss} + \underbrace{\lambda_D \cdot f_{obj} (o_i^{T}, \hat{o_i})}_\text{Distillation loss} 
\end{equation}
The objectness scaled classification function for the student network is given as: 
\begin{equation}
f_{cl}^{Comb} (p_i^{gt},\hat{p_i}, p_i^{T}, \hat{o_i^T} ) = f_{cl} (p_i^{gt}, \hat{p_i}) + \hat{o_i^T} \cdot \lambda_D \cdot f_{cl} (p_i^{T}, \hat{p_i})
\end{equation}
where the first part of the function corresponds to the original detection function while the second part is the objectness scaled distillation part. Following the similar idea the bounding box coordinates of the student network are also scaled using the objectness
\begin{equation}
f_{bb}^{Comb} (b_i^{gt},\hat{b_i}, b_i^{T}, \hat{o_i^T}) = f_{bb} (b_i^{gt}, \hat{b_i}) + \hat{o_i}^T \cdot \lambda_D \cdot f_{bb} (b_i^{T}, \hat{b_i}). 
\end{equation}
A large capacity teacher network assigns very low objectness values to a majority of the candidates which corresponds to the background.  The objectness based scaling act as a filter for distillation in single stage detector as it assigns a very low weight to the background cells. The foreground regions which appears like objects have higher values of objectness in the teacher network and the formulated distillation loss utilizes the teacher knowledge of these regions. It should be noted that the loss function stays the same but for distillation, we only add the teacher output instead of the ground truth. The loss function for the training is given as: 
\begin{multline}
L_{final} = f_{bb}^{Comb} (b_i^{gt},\hat{b_i}, b_i^{T}, \hat{o_i^T}) \\  + f_{cl}^{Comb} (p_i^{gt},\hat{p_i}, p_i^{T}, \hat{o_i^T} ) +  f_{obj}^{Comb} (o_i^{gt},\hat{o_i}, o_i^{T})
\end{multline}
which considers the detection and distillation loss for classification, bounding box regression and objectness.  It is minimized over all anchor boxes and all the cells locations of last convolutional feature maps.

\begin{figure}
\centering
   \includegraphics[trim = {4cm 6cm 7cm 6cm}, clip, width=\linewidth]{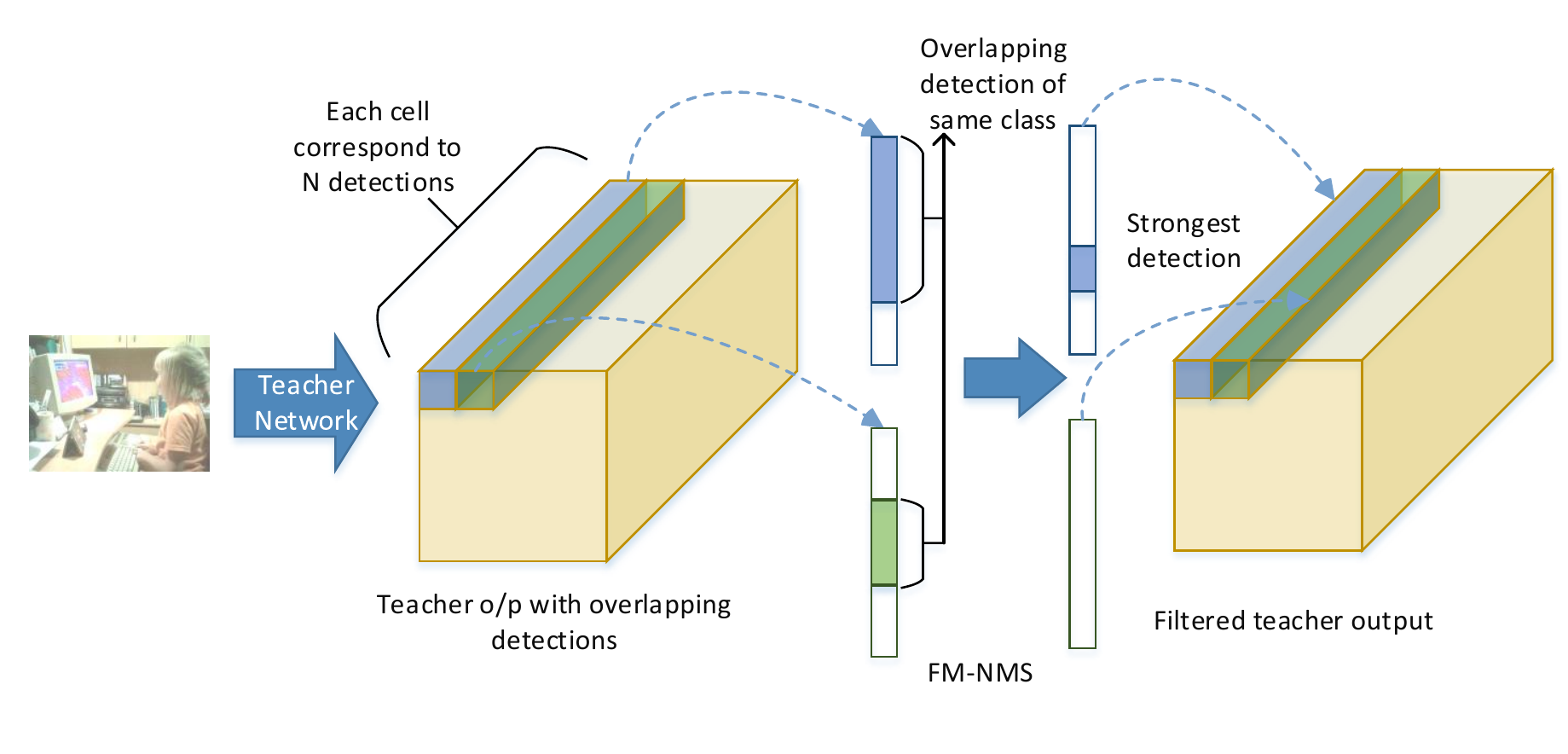}
  \caption{ Teacher network predicts bounding box coordinates and class probabilities simultaneously in the last layer. Each column represented in color blue and green corresponds to N detection, where N is number of anchor boxes. Adjacent columns often result in highly overlapping bounding boxes with the same class label. Proposed FM-NMS retains only the strongest candidate from the adjacent cells. This candidate is transfered as soft label to the student network. }
     \label{fig:distill-cells}
\end{figure}

\subsection{Feature Map-NMS}
Another challenge that we face comes from the inherent design of the single stage object detector. The network is trained to predict a box from a single anchor box of a cell, but in practice, a number of cells and anchor boxes end up predicting the same object in an image. Therefore, NMS is essential as a post-processing step in object detector architectures. However, the NMS step is applied outside the end-to-end network architecture and highly overlapping prediction are represented in the last convolutional layer of the object detector. When these predictions are transferred from the teacher to student network it results in a  redundant information.  Therefore, we observed that the distillation loss described above results in a loss of performance as the teacher network ends up transferring the information loss for highly overlapping detections. The feature maps corresponding to highly overlapping detection end up propagation large gradient for same object class and dimensions, thereby leading to network over-fitting. 

 In order to overcome the problem arising from overlapping detections, we propose Feature Map-NMS (FM-NMS). The idea behind FM-NMS is that if multiple candidates in neighbourhood of KxK cells correspond to the same class, then they are highly likely to correspond to the same object in an image. Thereby, we choose only one candidate with the highest objectness value. In practice, we check the activations corresponding to the class probabilities in last layer feature maps, set to zeros the activations correspond to the same class. The idea is demonstrated in Fig. \ref{fig:distillnms}, where we show the soft labels of teacher network in form of detections. The last layer of the teacher network predicts a number of bounding boxes in a region around the dog. To suppress the overlapping detections we pick the detection with the highest objectness values. The strongest candidate among the overlapping candidates is transfered to the student network. The idea for two cells is demonstrated in Fig. \ref{fig:distill-cells}. In our experiments we use the cell neighbourhood of $3 \times 3$.

\section{Effectiveness of data}
Finally in this paper we investigate the how much can we improve the performance by using more training data.  

\textbf{Labeled data} The straightforward approach is to add more annotated training data for the training. It has been shown \cite{zhu2016we,liu2016ssd} that increasing annotated data improves the performance of the models. However, earlier studies did not have a constraint of a model with limited capacity in their experiments. In this work, we restrict ourselves to a simple model and analyze if by simply adding the adding more data we can increase the performance. 

\textbf{Unlabeled data} Given the limited availability of annotated data, we utilize the unlabeled data in combination with the distillation loss. The main idea behind our approach is to use both soft labels and ground truth label when they are available. When the ground truth is not available only the soft labels of the teacher network are utilized. In practice, we propagate only the teacher part of the loss when the ground truth is not present and a combination of loss described in (2)-(4) otherwise. Since the objective function seamlessly integrate soft-label and ground truth, it allows us to train a network with a mix of the labeled and unlabeled data.

\section{Experiments on Object detection}
We perform experiments on Pascal VOC 2007 dataset \cite{everingham2010pascal}. The dataset consists of 20 object classes and 16K training images.

\subsection{Implementation Details}
We use Darknet deep learning framework \cite{redmon2016darknet} for our evaluation. Tiny-Darknet trained on the ImageNet \cite{russakovsky2015imagenet} for classification task is used for initialization. We remove the last two layers of the pre-trained network and add additional convolutional layers at the end. For detection, the network is trained using SGD with the initial learning rate of $10^{-3}$ for first 120 epochs and $10^{-4}$ for next 20 epochs and finally $10^{-5}$ for last 20 epochs. We use the standard training strategies such as momentum of 0.9 and 0.0005 weight decay. The batch size is set to 32 in all our experiments. The size of the input image in our experiments is set to $416 \times 416$. For the network distillation experiments we set the $\lambda_D$ to 1, thereby giving equal weight to the distillation and detection loss, however as the distillation part is scaled to the objectness, the final weight of the distillation part is always less than the detection loss.  

\begin{table}[t]
  \centering
 \resizebox{0.95\columnwidth}{!}{%
\begin{tabular}{| l | c | c | c | c | c | c | c | c |}
  \hline
  Conv2  &  Conv3   &  Conv4   &  Conv5 &  Conv6  &  Conv7    & Conv11          & max  & stack  \\ \hline
  \multicolumn{7}{|c|}{{Baseline}}  &                                           54.2 & 54.2   \\ \hline
   X  &  X  &  X   &  \checkmark &  \checkmark  &  X & X &                      55.8 & 56.7   \\ \hline
   X  &  X  &  \checkmark   &  \checkmark &  \checkmark  & X & X &              56.1 & 57.6   \\ \hline
   X  &  \checkmark  &  \checkmark   &  \checkmark &  \checkmark  &  X & X &    58.1 & 58.4   \\ \hline
  \checkmark  &  \checkmark   &  \checkmark   &  \checkmark &  \checkmark  & X& X&  57.7 & 58.2 \\ \hline
   X  &  \checkmark  &  \checkmark   &  \checkmark &  \checkmark  & \checkmark & X & 58.4& 58.9 \\ \hline
 X  &  \checkmark   &  \checkmark   &  \checkmark &  \checkmark  &   \checkmark & \checkmark & 58.6 & 59.4   \\ \hline
\end{tabular}
} 
\caption{\label{tab:baseline_arch} The accuracy of the detector after merging different layers. }
\end{table}

\begin{table}[t]
  \centering
  
 \resizebox{0.75\columnwidth}{!}{%
\begin{tabular}{| l | c |}
  \hline
  Modifications                    & Speed    \\ \hline
  Yolo-v2                          & 38   \\ \hline
  Tiny Yolo                        & 204       \\ \hline
  Merging multiple layers          & 184    \\ \hline
  Reducing the channels from 1024 to 512     & 234      \\ \hline
  Adding $1 \times 1$ convolution  & 221   \\ \hline
\end{tabular}
}
\caption{\label{tab:architecture-time} Speed comparison for different architecture modifications. }
\end{table}

\begin{table}[t]
  \centering
  
 \resizebox{0.9\columnwidth}{!}{%
\begin{tabular}{| l | c | c | c |}
  \hline
                                 &       &  \multicolumn{2}{|c|}{{Teacher Networks}}        \\ \hline
  Distill Config.                & Data  &  Pascal Teacher   &  COCO Teacher        \\ \hline
  Teacher Baseline               & -     &  73.4           &  76.9                  \\ \hline
  Student Baseline (No distill.) & VOC   &  59.4           &  59.6                  \\ \hline
  Distillation w/o FM-NMS        & VOC  &  57.1 (-2.3)     &  57.0 (-2.6)             \\ \hline
  Distillation w/o Obj-scaling   & VOC &  59.2  (-0.2)     &  59.1 (-0.5)        \\ \hline
  Distillation full              & VOC &  60.3 (+0.9)      &  60.4 (+0.8)              \\ \hline
  Student Baseline (No distill.) & VOCOCO   &  64.2        &  63.9                  \\ \hline
  Distillation w/o FM-NMS        & VOCOCO  &  61.1 (-3.1)  &  61.8 (-2.1)               \\ \hline
  Distillation w/o Obj-scaling   & VOCOCO &  65.8  (+1.2)  &  64.9 (+1.0)           \\ \hline
  Distillation full              & VOCOCO &  66.9  (+2.7)  &  66.1 (2.1)          \\ \hline
\end{tabular}
}
\caption{\label{tab:distill-config} Comparison of performance for distillation with different strategies on Pascal VOC 2007. The results are shown for two teacher network and for two set of labeled training data (Pascal VOC and combination of Pascal VOC and COCO).  }
\end{table}

\subsection{Architecture}
In this section, we present the results of different architecture configurations. First, we evaluate the effect of the feature map merging on the base architecture. The results for different layer combinations are shown in Table \ref{tab:baseline_arch}. It can be observed that the accuracy increases as the feature maps from more layers are merged together. Another important inference that we can draw from these results is that merging more advanced layers results in a more improvement rather than the initial layers of the network. There is a slight drop in the performance as the first few convolutional layers are merged with the final combination, indicating that the initial layer capture quite rudimentary information. Conv11 column of the table corresponds to the additional $1 \times 1$ convolutional layers added at the end to increase the depth of the network. These layers result in a gain of 0.5 mAP and provide a computationally efficient way of increasing the depth.

 We also show the comparison of two different approaches for feature map merging. Max layers were used in most of the prior works \cite{liu2016ssd, kim2016pvanet}, while feature stacking is a less common approach \cite{redmon2016yolo9000}. We found that feature stacking achieves much better results than max pooling. It can be observed for all combinations of merging the stacking achieves a better accuracy. 
 
Table \ref{tab:architecture-time} shows the speed of various improvement on the baseline detector. The speed is shown for a single Nvidia GTX 1080 GPU with 8 GPU memory and 16 GB CPU memory. The experiments are performed in batches and the small size of the network allows us to have a larger batch size and also enables parallel processing. Given that the GTX 1080 is not the most powerful GPU currently available, we believe that these models would be even faster on a more advanced GPU like Nvidia Titan X, etc. For the baseline Tiny-Yolo, we are able to achieve the speed of more than 200 FPS, as claimed by the original authors, using parallel processing and batch implementation of the original Darknet library. All the speed measurement are performed on Pascal VOC 2007 test image and we show the average time for 4952 images and it also includes the time for writing the detection results in the file. From the results, it can be observed that merging operations reduces the speed of the detector as it results in a layer with a fairly large number of feature maps. The convolutional operation on the combined feature maps layer reduces the speed of detector.  Therefore, it can be observed that reducing the number of feature maps has a big impact on the speed of the detector. We are able to push the speed beyond 200 by reducing the filter to 512 instead of 1024.  Finally, adding more $1 \times 1$ layers at the end of architecture also comes at fairly low computational overhead. These simple modifications result in an architecture which is an order of magnitude faster than popular architectures available.

\begin{table*}[t]
  \centering
  
 \resizebox{2\columnwidth}{!}{%
\begin{tabular}{| l | c | c | c  c  c  c  c  c  c  c  c  c  c  c  c  c  c  c  c  c  c  c |}
  \hline
  Algo. &  Info & mAP    & Aero & bike & bird & boat & bottle & bus & car & cat & chair & cow & table & dog & horse & mbike & person & plant & sheep & sofa & train & tv  \\ \hline

   Yolo & - & 57.9  & 77.0 & 67.2 & 57.7 & 38.3 & 22.7 & 68.3 & 55.9 & 81.4 & 36.2 & 60.8 & 48.5 & 77.2 & 72.3 & 71.3 & 63.5 & 28.9 & 52.2 & 54.8 & 73.9 & 50.8  \\ \hline
   
   Yolo-v2 & Teacher & 76.8  &    80.6  &   83.9  &   75.2  &   61.1 &   54.3  &   86.5 &    81.1   & 89.0 &    60.2  &   85.4 &   71.2 &     86.7   &   89.2 &   85.7   &   77.9  &    50.3   &    79.4   &  75.9   &    86.3  &    77.2  \\ \hline

  Tiny-Yolo & Baseline & 54.2 &     57.4  &  67.5  &  44.9    & 34.8 &   20.4  &   67.5   &  62.9  &   67.4    &  32.0  & 53.7  &    58.1  &    61.6  &    70.5  &   69.1 &   58.0 &  27.8  &  52.8  &  51.1 & 68.5 &  57.4  \\ \hline

    F-Yolo & Arch Modified & 59.4 &     61.6  &  71.3  &  49.4 &    43.0  &  29.4 &    70.0  &  68.7 &  70.1 &  38.4   & 59.3   & 60.9  &  68.1  &  73.8  &   72.6  &     64.4  &  32.2 &     60.0  &   59.2 &   75.1  &   61.2  \\ \hline
    
    D-Yolo &  Distiled  & 66.9  & 69.6    & 77.1   & 59.6   & 49.6  & 39.0  &  76.9    & 74.2   & 78.8    &  45.8  & 71.0   & 69.3   & 72.4    & 81.5   & 77.9    & 72.4   & 40.0   & 68.5   & 67.0   &  78.2   &  68.0 \\ \hline
\end{tabular}
}
\caption{\label{tab:pascal-comparison} Comparison of proposed approach with the popular object detectors on VOC-07 dataset. }
\end{table*}
 
 
\subsection{Distillation with labeled data}
 First, we describe our teacher and student networks. Since we use the soft labels of the teacher network for the training, we are confined to use a Yolo based teacher network which densely predicts the detections as the output of last convolutional layer. Another constraint on the selection of the teacher/student network is that they should have same input image resolution because a transfer of the soft-labels requires the feature maps of the same size. Thereby, we choose Yolo-v2 with Darknet-19 base architecture as the teacher. We use the proposed F-Yolo as the student network, as it is computationally light-weight and is very fast. To study the impact of more labelled data we perform training with only Pascal data and combination of Pascal and COCO data. For Pascal we consider the training and validation set from 2007 and 2012 challenge and for COCO we select the training images which have at least one object of Pascal category in it. We found there are 65K such images in COCO training dataset.

 To study the impact of teacher network on the distillation training we also train our teacher models with two different datasets:  Pascal data and combination of COCO and Pascal. The baseline performance of these teacher networks is given in Table \ref{tab:distill-config}. It can be observed that simply by training Yolo-v2 with COCO training data improves the performance by 3.5 points. With these two teachers, we would like to understand the effect of a more accurate teacher on the student performance. 
 
In our first experiment, we compare different strategies to justify the effectiveness of our proposed approach. We introduce two main innovation for single stage detector: Objectness scaling and FM-NMS. We perform distillation without the FM-NMS and objectness scaling step. The results are shown in Table\ref{tab:distill-config}. It can be observed that the performance of the distilled student detector drops below the baseline detector when distillation is performed without FM-NMS step. For both the teacher network there is a significant drop in the performance of the network.  Based on these experiments we find that the FM-NMS is a crucial element to make distillation work on single stage detector. In the experiments without the objectness scaling, we again observe a drop in the performance, although the drop in the performance is not very high. 

 
 The experiments with additional annotated data (COCO training image) show a similar trend, thus verifying the importance of FM-NMS and object scaling. However, it is interesting to observe that there is a significant improvement in the performance of full distillation experiment with larger training data. Full distillation approach gain by 2.7 mAP with COCO training dataset. With larger training data there are soft-labels which can capture much more information about the object like section present in the image. 
 
 We can also observe that the performance of the baseline detector improves significantly with larger training data.  It shows that our light-weight models have the capacity to learn more provided more training samples. With distillation loss and additional annotated data proposed detector achieves 67 mAP while running at a speed of more than 200 FPS. 
 
 Surprisingly, for a fixed student the teacher detector does not plays a crucial role in the training. The COCO teacher performs worse than the VOC teacher when combination of VOC and COCO data is used. We suspect the it is difficult to evaluate the impact of the teacher quality as the different in the teacher performance is not large (< 4mAP). 
 
\begin{figure*}[!tbp]
    \centering
    \begin{subfigure}[b]{0.3\textwidth}
        \includegraphics[width=\textwidth]{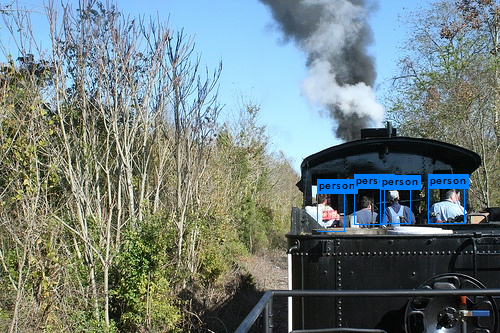}

    \end{subfigure}
    ~ 
    \begin{subfigure}[b]{0.3\textwidth}
        \includegraphics[width=\textwidth]{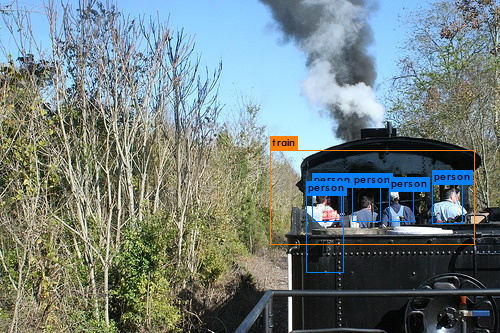}

    \end{subfigure}
    ~ 
    \begin{subfigure}[b]{0.3\textwidth}
        \includegraphics[width=\textwidth]{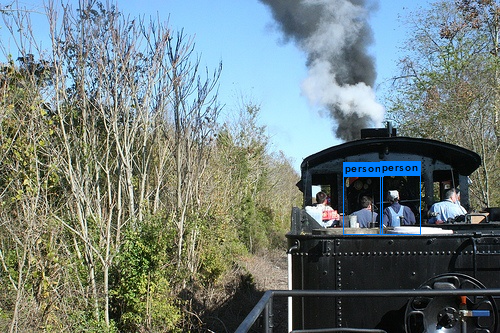}

    \end{subfigure}
  
     ~ 
    \begin{subfigure}[b]{0.3\textwidth}
        \includegraphics[width=\textwidth]{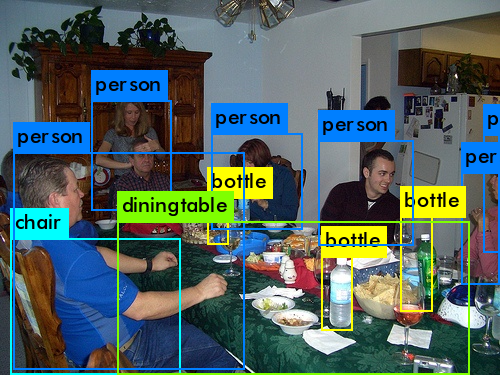}

    \end{subfigure}
    ~ 
    \begin{subfigure}[b]{0.3\textwidth}
        \includegraphics[width=\textwidth]{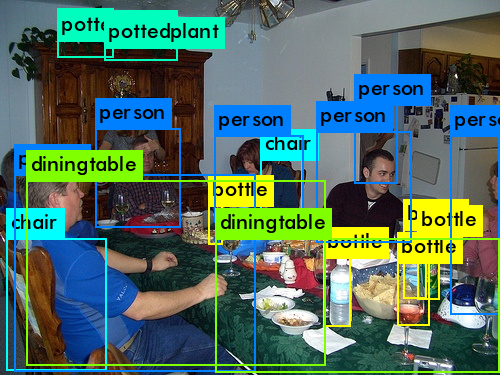}

    \end{subfigure}
     ~\begin{subfigure}[b]{0.3\textwidth}
        \includegraphics[width=\textwidth]{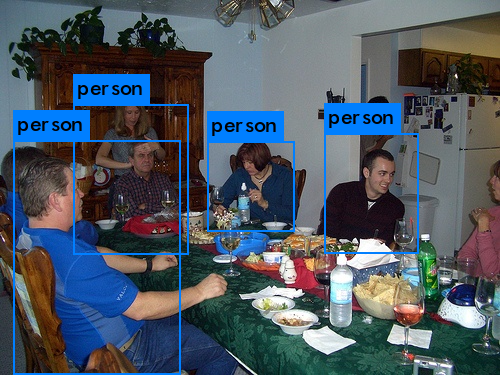}

    \end{subfigure}
     ~ 
    \begin{subfigure}[b]{0.3\textwidth}
        \includegraphics[width=\textwidth]{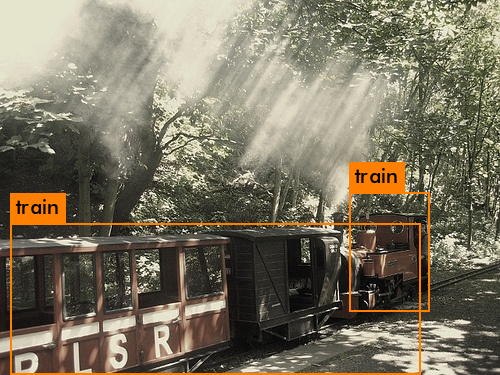}
       \caption{Yolo-v2 Teacher}
    \end{subfigure}
    ~ 
    \begin{subfigure}[b]{0.3\textwidth}
        \includegraphics[width=\textwidth]{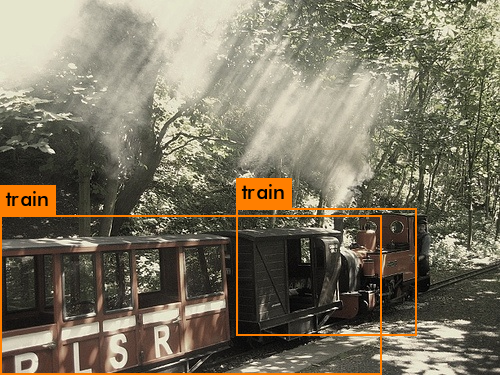}
        \caption{Distilled Yolo}
    \end{subfigure}
     ~\begin{subfigure}[b]{0.3\textwidth}
        \includegraphics[width=\textwidth]{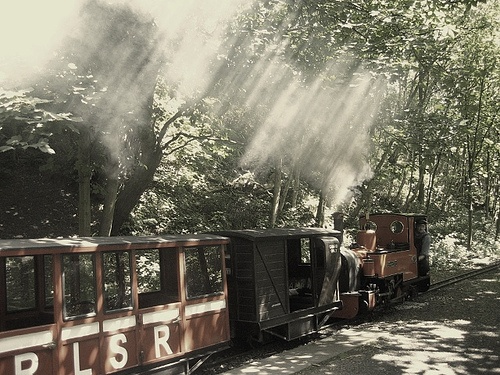}
        \caption{Tiny-Yolo}
    \end{subfigure}
    \caption{Example images with teacher network (Yolo-v2), proposed approach and the Tiny-Yolo baseline. } \label{fig:detection-comparison}
\end{figure*}

 We show the detectors performance for the different classes of the Pascal VOC 2007 test set in Table \ref{tab:pascal-comparison}. The performance of the proposed F-Yolo (only with architecture modifications) and D-Yolo (architecture changes + distillation loss) is compared with original Yolo and Yolov2. It is interesting to observe that with distillation loss and more data there is a significant improvement for small objects such as bottle and bird (10 AP). The difference in the performance between the Tiny-Yolo and the proposed approach is clearly approach in some of the sample images shown in Fig. \ref{fig:detection-comparison}.

\begin{table}[t]
  \centering
  
 \resizebox{0.9\columnwidth}{!}{%
\begin{tabular}{| l | c | c | c | c | }
  \hline
  Loss function              & Labeled  &  Unlabeled     &  VOC Teacher   \\ \hline
  No distillation            & VOC     &     -            &  59.4        \\ \hline
  Distillation               & VOC     &   -              &  60.3       \\ \hline
  Distillation               & VOC     &  COCO-16k       &  61.5        \\ \hline
  Distillation               & VOC     &  COCO-32K       &  61.8        \\ \hline
  Distillation               & VOC     &  COCO-48K       &  62.1        \\ \hline  
  Distillation               & VOC     &  COCO-65K       &  62.3        \\ \hline
\end{tabular}
}
\caption{\label{tab:distill-unlabeled} Performance comparison with on Pascal 2007 using unlabeled data. }
\end{table} 
 
\subsection{Unlabeled data}
 Previous experiment with combination of COCO and VOC data showed that the F-Yolo has the capacity to learn with more training data. In this section we study how much can our model learn from unlabeled data. In this experiment we evaluate accuracy of the detector by increasing the unlabeled data in the training set. We use labeled data only from VOC dataset and use the images from COCO without their labels as additional data. The labeled and unlabeled images are combined and used together for training, for unlabeled images we only propagate the loss evaluated from teacher soft-labels. We train the network with different number of unlabeled images (16K, 32K, 48K and 65K) to evaluate the influence of the unlabeled data on the student network. The results are shown in Table \ref{tab:distill-unlabeled}. It can be observed that the performance of the baseline detector improves significantly (3-4 mAP) with additional data. As more unlabeled data is added the performance of the detector improves. 
 
 It is interesting to compare the change in the performance with unlabeled data and COCO labeled data separately to understand the importance of annotation in the training. Using complete COCO data with annotation our model achieve 64.2 mAP ( Table \ref{tab:distill-config} student baseline) and using Yolo-v2 as teacher network and unlabeled COCO images, model obtain 62.3 mAP. These results indicate that although the annotation are important, we can significantly improve the performance by using an accurate teacher network simply by adding more unlabeled training data.

 Finally, we compare the performance of proposed distillation approach with the competing distillation approach \cite{chen2017learning}. The competing approach for distillation employs Faster-RCNN framework which is a two stage detector, while our approach has been specifically designed for one stage detector.  The results are shown in Table \ref{tab:distill-rcnn-comparison}. The performance is compared with the following architectures: Alexnet \cite{krizhevsky2012imagenet}, Tuckernet \cite{kim2015compression}, VGG-M \cite{simonyan2014very} which are distilled using VGG-16 network \cite{simonyan2014very}. It can be observed that the proposed distilled network is an order of magnitude faster than the RCNN based detector. In terms of number of parameters the proposed approach is comparable to Tucker network, however, it is much faster than all Faster-RCNN based networks shown here. The speed-up over these approaches can be attributed to the efficient design of single stage detector and the underlying network optimized for speed and the additional data that is used for training. The results show that these modifications leads to a gain of around 9 mAP over the competing comparable network while being much faster than it.

\begin{table}[t]
  \centering
  
 \resizebox{0.9\columnwidth}{!}{%
\begin{tabular}{| l | c | c | c | c |}
  \hline
         
  Network    & Framework & Speed  &  mAP   &   Params (million)       \\ \hline
  Tucker     & Faster-RCNN & 21   &  59.4   &   11M                      \\ \hline
  Alex       & Faster-RCNN & 13   &  60.1   &   62M                 \\ \hline
  VGG-M      & Faster-RCNN & 13   &  63.7   &   80M                 \\ \hline
  D-Yolo     & Yolo        & 207  &  67.6   &   15M                 \\ \hline

\end{tabular}
}
\caption{\label{tab:distill-rcnn-comparison} Comparison of proposed single stage distilled detector (Yolo) with Faster-RCNN distilled detectors on VOC 2007 test set.  }
\end{table}

 
 
 \section{Conclusions}
 In this paper, we develop an architecture for efficient and fast object detection.  We investigate the role of network architecture, loss function and training data to balance the speed performance trade-off. For network design, based on prior work, we identify some simple ideas to maintain computational simplicity and following up on these ideas we develop a light-weight network. For training, we show distillation is a powerful idea and with carefully designed  components (FM-NMS and objectness scaled loss),  it improves the performance of a light-weight single stage object detector. Finally, building on distillation loss we explore unlabeled data for training. Our experiments demonstrate the design principle proposed in this paper can be used to develop object detector which is an order of magnitude faster than the state-of-the-art object detector while achieving a reasonable performance. 


{\small
\bibliographystyle{ieee}
\bibliography{egbib}
}

\end{document}